\documentclass[journal]{IEEEtran} 
\usepackage{cite}
\usepackage{amsfonts,amssymb} 
\usepackage{url}
\usepackage{amsmath}
\usepackage{graphicx} 
\graphicspath{{./Fig/}}

\ifCLASSINFOpdf
\else
\fi
\begin{document} 

%
\title{A Review of Mobile Robot Motion Planning Methods: from Classical Motion Planning Workflows to Reinforcement Learning-based Architectures}
%
%
%

\author{Lu Dong,~\IEEEmembership{Member,~IEEE,} 
		Zichen He, 
		Chunwei Song, 
		and Changyin Sun,~\IEEEmembership{Senior Member, ~IEEE}
%
\thanks{
L.Dong is with the School of Cyber Science and Engineering, Southeast University, Nanjing 211189, China (e-mail:ldong90@seu.edu.cn). 

Z.He and C.Song are with the College of Electronics and Information Engineering, Tongji University, Shanghai 201804, China (e-mail:1910646@tongji.edu.cn; 2030739@tongji.edu.cn).

C.Sun is with the School of Automation, Southeast University, Nanjing 210096, China, and also with the College of Electronics and Information Engineering, Tongji University, Shanghai 201804, China (e-mail: cysun@seu.edu.cn).
}
}

\maketitle

\begin{abstract}
Motion planning is critical to realize the autonomous operation of mobile robots. As the complexity and randomness of robot application scenarios increase, the planning capability of the classical hierarchical motion planners is challenged. With the development of machine learning, deep reinforcement learning (DRL)-based motion planner has gradually become a research hotspot due to its several advantageous feature. DRL-based motion planner is model-free and does not rely on the prior structured map. Most importantly, DRL-based motion planner achieves the unification of the global planner and the local planner. In this paper, we provide a systematic review of various motion planning methods. First, we summarize the representative and state-of-the-art works for each submodule of the classical motion planning architecture and analyze their performance features. Subsequently, we concentrate on summarizing RL-based motion planning approaches, including motion planners combined with RL improvements, map-free RL-based motion planners, and multi-robot cooperative planning methods. Last but not least, we analyze the urgent challenges faced by these mainstream RL-based motion planners in detail, review some state-of-the-art works for these issues, and propose suggestions for future research. 
\end{abstract}

\begin{IEEEkeywords}
Mobile robot, Reinforcement learning, Motion planning, Multi-robot cooperative planning.
\end{IEEEkeywords}

%
\IEEEpeerreviewmaketitle

\begin{figure*}[htb]
	\centering
	\includegraphics[width=6.3 in]{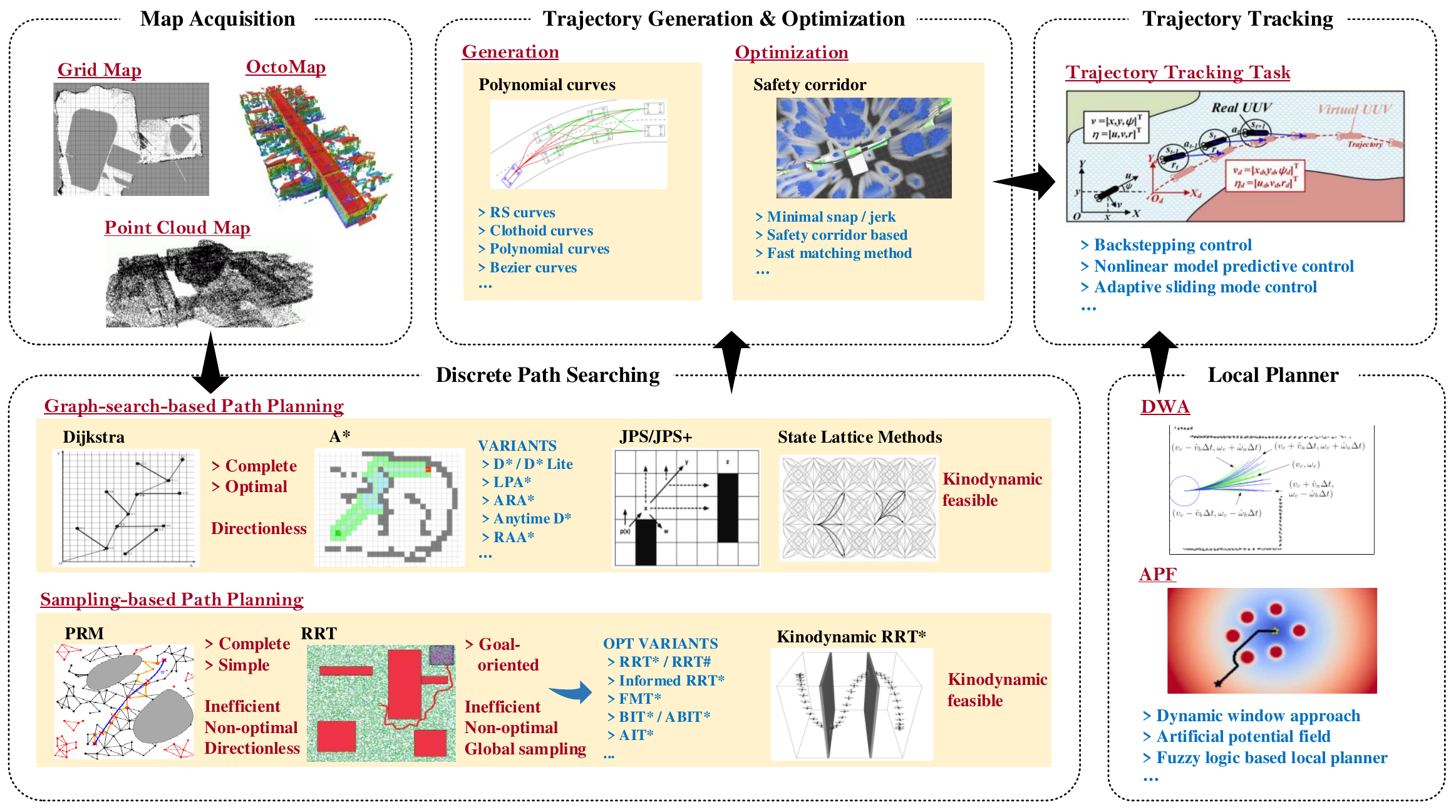}
	\caption{An overall architecture diagram of the classical motion planning workflow.}
	\label{Fig1}
\end{figure*}

\section{Introduction} 

%
%
%
%
\IEEEPARstart{W}{ith} the rapid development of AI, autonomous intelligent mobile robots (MRs) are always at the forefront of scientific research due to their compact size, flexible mobility, diverse functions, and modularity. MR can replace human beings to perform complicated and dangerous missions on various occasions by carrying different sensing modules. Therefore, it plays a vital role in ocean exploration, urban rescue, security patrol, and epidemic control \cite{xiao2020motion,9119863,aradi2020survey}, etc. 

The motion planning (MP) technology is one of the most critical modules that endows the MR with autonomous capabilities. The specific function of the motion planner is to integrate local or global state information of the robot system and produce optimal or near-optimal planning decisions in the face of various environments. The standard MP module consists of the global motion planner and the local motion planner. The global motion planner is responsible for generating optimal or near-optimal, kinodynamic feasible, safe, and executable trajectories on the basis of the structured prior map information. The local motion planner is responsible for helping the MR make real-time motion decisions in local dynamic environments (e.g., pedestrian participation environment, outdoor environments, etc.). 


The standard map-based MP framework is hierarchical and multi-level cascading and has a certain degree of customization. In the framework, the global planner and the local planner are independent of each other, and both need to be configured separately for different scenarios. These features make it difficult for classical motion planners to adapt to unstructured, complex, and dynamic environments. Therefore, studying a map-less motion planner with a certain generalization, robustness, and adaptability is of great significance. With the development of reinforcement learning (RL), RL-based motion planners can be independent of the map prior data and obtain a stronger generalization by learning interactively with various scenarios during the training stage. Therefore, it gradually becomes a research hotspot.



The research of RL has experienced a long history. The dynamic programming algorithm proposed by Bellman in 1956 has laid the foundation for the subsequent development of this field \cite{arulkumaran2017deep}. The primary research issue of the RL is the tradeoff between exploration and exploitation at each time step. The agent explores to discover different policies that can bring more incredible benefits or exploits current optimal policies. Along with the substantial improvement in the computing power and the storage capacity of hardware systems, deep learning (DL) technology has been widely used in RL. Researchers utilize DRL to integrate the learning module and the decision module, and realize the nonlinear mapping procedure from the raw state inputs to the final action decisions. With its superior properties, DRL has been successfully applied to gaming AI, autopilot, transportation schedules, power system optimization \cite{7556990,aradi2020survey,dong2020event}, etc. In the field of mobile robot MP, RL-based motion planners achieve end-to-end planning, get rid of the tedious hierarchical multi-level coupling planning framework, and unify the global motion planner with the local motion planner. By constructing specific reward forms and training paradigms according to different task objectives, the state updating policy can be improved iteratively based on the feedback signals from the environment during the process of interaction. These features mentioned above offer the possibility of deploying RL-based motion planners to robots operating in some unstructured and dynamic environments where the real-time mapping process is challenging to perform. 

This paper is a systematic review of current mainstream and state-of-the-art mobile robot motion planning methods. Its content mainly covers robot types, including wheeled mobile robots (WMRs), autonomous underwater vehicles (AUVs), unmanned aerial vehicles (UAVs), etc. The overall structure of this paper consists of three parts. The first part is the summary and comparison of different representative algorithms of each sub-module in the pipeline of the classical motion planner. The second part is an overview of RL-based MP approaches. It consists of three sections. The first section is a summary of classical motion planning methods incorporating the RL optimization module. This type of motion planner still relies on the map prior. The role of the RL module is to make local planning decisions (e.g., obstacle avoidance) or select the optimal functional hyper-parameters in classical motion planners. The second section is an overview of sensor-level end-to-end RL-based motion planners. This type of motion planner is map-free. We mainly review two mainstream sensor-based works: lidar-based methods and vision sensor-based methods. The third section is an overview of RL-based multi-robot collaborative motion planning methods. In this section, we focus on reviewing some works of multi-robot collaborative planning based on centralized training with decentralized execution (CTDE) RL paradigm. The last part of this survey is a discussion. In this part, we systematically summarize several challenges faced by current RL-based motion planners, which are reality gap, social etiquette, catastrophic forgetting problem, reward sparsity issue, lidar data pre-processing problem, low sample efficiency problem, and generalization problem. These issues hinder the application and deployment of the pretrained RL agents in realistic physical environment. Also, we review several representative works aimed at addressing these issues.

To sum up, the rest of this paper is organized as follows. Section II is a review of classical hierarchical MP approaches. Section III focuses on summarizing several map-based classical motion planners combined with the RL optimization algorithms. Section IV discusses the map-less and end-to-end RL-based MP methods at the sensor level. Section V provides a survey of RL-based multi-robot motion planning methods. Section VI concludes several current challenges in RL-based motion planners and gives future directions. The conclusion of this paper is drawn in section VII.

\section{Classical Motion Planning of MRs}   

Before we start designing the workflow of the classical motion planner, we need to obtain the map representation of the environment. Commonly used structured maps include occupancy grid map, point cloud map, Voronoi diagram map, Euclidean signed distance fields, etc \cite{oleynikova2017voxblox,han2019fiesta}. The quality and accuracy of these prior maps directly determine the final planning performance. We represent the classical hierarchical architecture of the classical motion planner as shown in Fig. \ref{Fig1}. This framework can be divided into four submodules: discrete path searching, trajectory generation and optimization, trajectory tracking, and local planner \cite{quan2020survey,claussmann2019review,xiao2020motion}. It can be found that the classical motion planning approach is map-based and highly customized for different mission scenarios. Each internal submodule is interdependent. In this section, we will describe the main features and principles of each submodule, list representative algorithms and their limitations, and provide an overview of recent progress. 

\subsection{Discrete Path Searching}
The goal of discrete path searching (DPS) process is to find a feasible path that consists of a series of discrete waypoints from the initial point to the target point. DPS works in the configuration space (C-space). Each configuration of the robot can be represented as a point. This approximation process reduces the complexity of the computation and improves search efficiency. Notably, in C-space, specific expansion operations are required for robots of different sizes and shapes \cite{choset2005principles}.  

Traditional global DPS algorithms can be divided into two categories: the graph-searching-based algorithm (GSBA) and the sampling-based algorithm (SBA) \cite{7339478}.  

\subsubsection{Graph-search-based algorithms}  
Depth-first search (DFS) and breadth-first search (BFS) are two fundamental graph search algorithms. On the basis of BFS, Dijkstra is proposed. This algorithm is greedy, complete, and optimal \cite{5987118}. However, Dijkstra lacks directionality in the process of path search. In \cite{hart1968formal}, A* is proposed. The researchers introduce the heuristic function to measure the distance between the real-time search position and the target position. This function makes the search more oriented and improves the search speed compared to Dijkstra. In \cite{stentz1997optimal}, Anthony Stentz presents Dynamic A* (D*). They replace the heuristic rule in A* with an incremental reverse rule. Based on A*, SvenKoenig et al. develop Lifelong planning A* (LPA*) in \cite{koenig2004lifelong}. They combine incremental search with A*. LPA* avoids the problem of recalculating the whole graph due to changes in the environment. In \cite{belanova2018path}, Dorota Belanová et al. propose D* Lite. D* Lite is a path planning algorithm with the variable start point and the fixed target point. This method incorporate the reverse searchint trick with the heuristic mechanism. The difference between D* Lite and LPA* is the search direction. Jump point search (JPS) \cite{harabor2011online} is another type of GSBA with different search principles. The proposers seek to improve the search efficiency by optimizing subsequent nodes searching process on the basis of A*. Notably, JPS only adds the jump points that are searched according to the specific rules into the open list. This operation excludes a large number of meaningless nodes. Therefore, JPS occupies less memory and can search faster than A*. However, JPS is only applicable to the uniform grid map \cite{quan2020survey,9354210}. In \cite{9327403}, Changjiang Jiang et al. propose the JPS+ based path planning method. They further improve the search efficiency of JPS by adding the pre-processing section but is less suitable for dynamic environments. 


Some researchers focus on improving the real-time performance of A*, such as \cite{bulitko2006learning,koenig2006real}. However, none of the above methods consider the kino-dynamics of the robot. For some WMRs with non-holonomic constraints, sometimes the discrete path planned by the above methods cannot be executed well. State lattice methods \cite{pivtoraiko2012generating,6094900,zhou2019robust} have been proposed to handle this problem. These approaches perform spatial discretization, and use a hyper-dimensional grid of states to represent the planning area. The sampling process allows the planner to generate a series of dynamically feasible motion connections. Finally, the algorithm would search for the optimal path fragment among these connections.

\subsubsection{Sampling-based algorithms}{
GSBAs are mainly applied to path planning problems on low-dimensional spaces. The completeness of these methods depends on the entire modeling process of the environment. In a higher-dimensional space, such methods would be suffering from the curse of dimensionality. Sampling-based algorithm (SBA) are more suitable for high-dimensional path searching scenarios. These algorithms have probabilistic completeness and further improve the search efficiency of feasible waypoints \cite{gonzalez2015review}. 

Probabilistic road map (PRM) and rapid-exploring random tree (RRT) are two fundamental types of SBAs \cite{quan2020survey}. PRM builds a graph during the learning stage and utilizes it to search for valid discrete paths during the query stage \cite{9354210}. It is simple with few parameters but lacks optimality. RRT is more goal-oriented than PRM. It generates an extended tree by selecting leaf nodes with random sampling. When the leaf node expands to the target region, the discrete path from the root node to the goal is obtained. RRT is not sufficient because of the whole-space-sampling process. And, RRT is not an optimal or asymptotically optimal algorithm. Limited by these restrictions, RRT cannot plan a feasible path quickly in the narrow passage environment. Until now, there are still many researchers dedicated to optimizing these problems in RRT \cite{gonzalez2015review}. RRT* \cite{karaman2011sampling} introduces prune optimization and random geometric graphs (RGG) during the node extension phase of RRT. It is an asymptotically optimal algorithm. In \cite{nasir2013rrt,arslan2013use}, RRT*-smart and RRT\# are proposed respectively. The main idea of these two methods is to improve the convergence speed of RRT*. In \cite{6631299}, kino-dynamic RRT* is presented. This method additionally deals with the kino-dynamics of robots. Also, kino-dynamic RRT* samples in full state space and can handle the problem of robot motion planning with non-holonomic constraints. Informed RRT* in \cite{6942976} directly limits the sampling interval to improve the overall convergence efficiency, which can effectively solve the discrete path search problem for narrow passages. Batch informed trees (BIT*) is presented in \cite{gammell2015batch}. This method unifies the advantages of GSBAs and SBAs and introduces a heuristic method to search for a sequence of increasingly dense implicit RGGs iteratively. Compared with RRT*, informed-RRT*, and fast match trees (FMT*) \cite{janson2015fast}, BIT* enhances planning performance significantly in the experiments. In \cite{strub2020advanced}, Marlin P. Strub et al. further extend BIT* utilize advanced truncated anytime graph-based search techniques to enhance real-time planning performance. In \cite{strub2020adaptively}, they introduce asymmetric bidirectional search on the basis of BIT*. This trick can help the planning process converge towards the optimal result as fast as RRT-connect. Real-time RRT* (RT-RRT*) \cite{naderi2015rt} and information-driven RRT* (ID-RRT*) \cite{pimentel2018information} focus on enhancing the real-time performance of the RRT*, improving the planning capability in unknown and dynamic environments.}

\subsection{Trajectory Generation and Optimization} 
Most of the DPS algorithms only consider the geometric constraints of the workspace. In some cases, the final optimal or near-optimal segmented trajectories are not executable for the actual MR. The trajectory generation and optimization (TGO) process considers multiple constraints (e.g., safety constraints, kino-dynamic constraints, etc.) of the robot and incorporates the time-allocation mechanism in the planning process to empower the motion planner with more executable capabilities. By cascading with TGO, the planner can finally generate a kino-dynamic feasible, collision-free, executable, and trackable trajectory that satisfies several optimization objectives (time optimal, energy optimal, etc.).

The interpolation-curve-based method is one of the most commonly used approaches for trajectory generation. This method can generate trajectories with good continuity and differentiability. The typical interpolating curves include Reeds and Sheep (RS) curves \cite{fraichard2004reeds}, clothoid curves \cite{brezak2013real}, polynomial curves \cite{gonzalez2015review}, Bezier curves \cite{gao2018online}, etc. The minimum snap presented in \cite{mellinger2011minimum} is an effective optimization paradigm for trajectory optimization and has inspired many scholars. The method proposer Mellinger utilizes the differential flatness of the drone to reduce the dimension of the state space and the action space. By solving the quadratic programming (QP) problem with several constraints, the minimal snap algorithm can minimize the thrust change rate to achieve the objective of optimal energy consumption.  Later, Richter et al. solved the minimum snap problem in closed-form to avoid the numerical instability \cite{richter2016polynomial}. In \cite{chen2016online}, Chen Jing et al. introduce safety corridor constraints in the minimum snap algorithm to enforce the safety of the robot during planning. However, the process of iterative detection of the boundary extremum safety is time-consuming. Fei Gao et al. use Bezier curves with the features of convex hull and hodograph to substitute traditional polynomial curves \cite{gao2018online}. This approach has fewer constraints and avoids the tedious iterative checking process. In \cite{gao2017gradient}, Fei Gao et al. present an online TGO method based on the Euclidean distance field (EDF). The cost function of this method is composed of the elastic band smoothness term, the safety term, and the kino-dynamic term. Then, the nonlinear optimization method is used to solve the final trajectory. EDF-based TGO method is real-time, and has great local replanning ability. It overcomes the problem of low clearance between the MR and the obstacle in previous approaches. 
 
\subsection{Trajectory Tracking} 
Generally, the purpose of the trajectory tracking (TT) phase is to enable the MR to track the trajectory planned by the TGO process. The early TT task belongs to the control level. It mainly focuses on designing virtual controllers based on dynamics equations so that the MR can track a given reference trajectory asymptotically. Common trajectory tracking control methods are input-output linearization \cite{8945483}, backstepping control \cite{sun2013bioinspired}, sliding mode control (SMC) \cite{ibrahim2016wheeled}, robust control \cite{osusky2018trajectory}, etc. However, these approaches suffer from several limitations. For example, the mathematical models of some robots (e.g., AUVs) contain complex nonlinear or uncertain terms. These terms increase the difficulty of modeling. Besides, the realistic operating environments of MR are often changeable, such as crowded environments, multi-robot interaction environments. These challenges require that the TT algorithms should have certain anti-disturbance and local replanning capabilities. 

In \cite{8447570}, an adaptive sliding mode controller (ASMC) is designed to handle the TT problem of the WMRs. This algorithm takes nonlinear model and disturbances into account and utilizes the discontinuous projection mapping to adjust the performance parameters of the controller. Model predictive control (MPC) is also a mainstream approach used to deal with TT problems. MPC belongs to the category of optimal online  control method and can handle various state and control constraints \cite{nascimento2018nonholonomic}. In \cite{7272877}, the MPC-based iterative trajectory tracking scheme is presented and applied to UAV navigation. In \cite{9257897}, Avraiem Iskander et al. select to adopt the nonlinear MPC (NMPC) to cooperate with RRT* and the minimum snap algorithm to build a closed-loop of UAV motion planning in three-dimensional (3D) space. Björn Lindqvist et al. focus on coping with dynamic obstacle avoidance problems. They couple the dynamic obstacle avoidance strategies in the TT controller. The proposed architecture of the novel NMPC is based on the PANOC non-convex solver and the trajectory classification scheme \cite{9145644}. Caicha Cui et al. combine the  MPC with the robust sliding mode dynamic control (RSMDC). The MPC module is responsible for replanning trajectory to avoid local obstacles. The RSMDC plays the role of controlling the tracking speed to reduce the impact of UUV model uncertainty and external disturbance on the final planning effect \cite{8407449}. 

\subsection{Local Planning} 
In the real world, MRs are always deployed in environments full of uncertain factors. For example, serving robots in airports or stations must reasonably predict and react to pedestrians. Therefore, the MR needs to have the local replanning ability to deal with unexpected situations while tracking the global trajectory. Local motion planners serve this purpose. 


The commonly used local planners include artificial potential field \cite{di2020local}, reactive replanning method \cite{minguez2000nearness,seder2007dynamic}, fuzzy algorithm based method \cite{chen2011mobile}, etc. APF is relatively simple and has good real-time planning performance. However, the traditional APF method is prone to fall into local optimum and has accessibility problem when the target point is surrounded by obstacles. Therefore, many researchers pay their attention to solving the shortcomings of the traditional APF. In \cite{di2020local}, Di Wang et al. solve the above limitations by introducing the left tuning potential field and the virtual target point. 

Reactive replanning methods can also avoid unknown dynamic obstacles in the environment. Common methods include directional approach \cite{minguez2000nearness}, dynamic window approach (DWA) \cite{seder2007dynamic}, etc. DWA is an active velocity selection algorithm and considers the kino-dynamics of the robot. DWA samples multiple velocities in the velocity space and generates a series of intrinsic motion trajectories in a certain time. By comparing the scores of different trajectories, the algorithm would select the optimal trajectory for the robot. DWA is usually coupled with the global trajectory tracking process and can endow the MR with a high degree of flexibility. 

\begin{figure*}[!t]
	\centering
	\includegraphics[width=6.2 in]{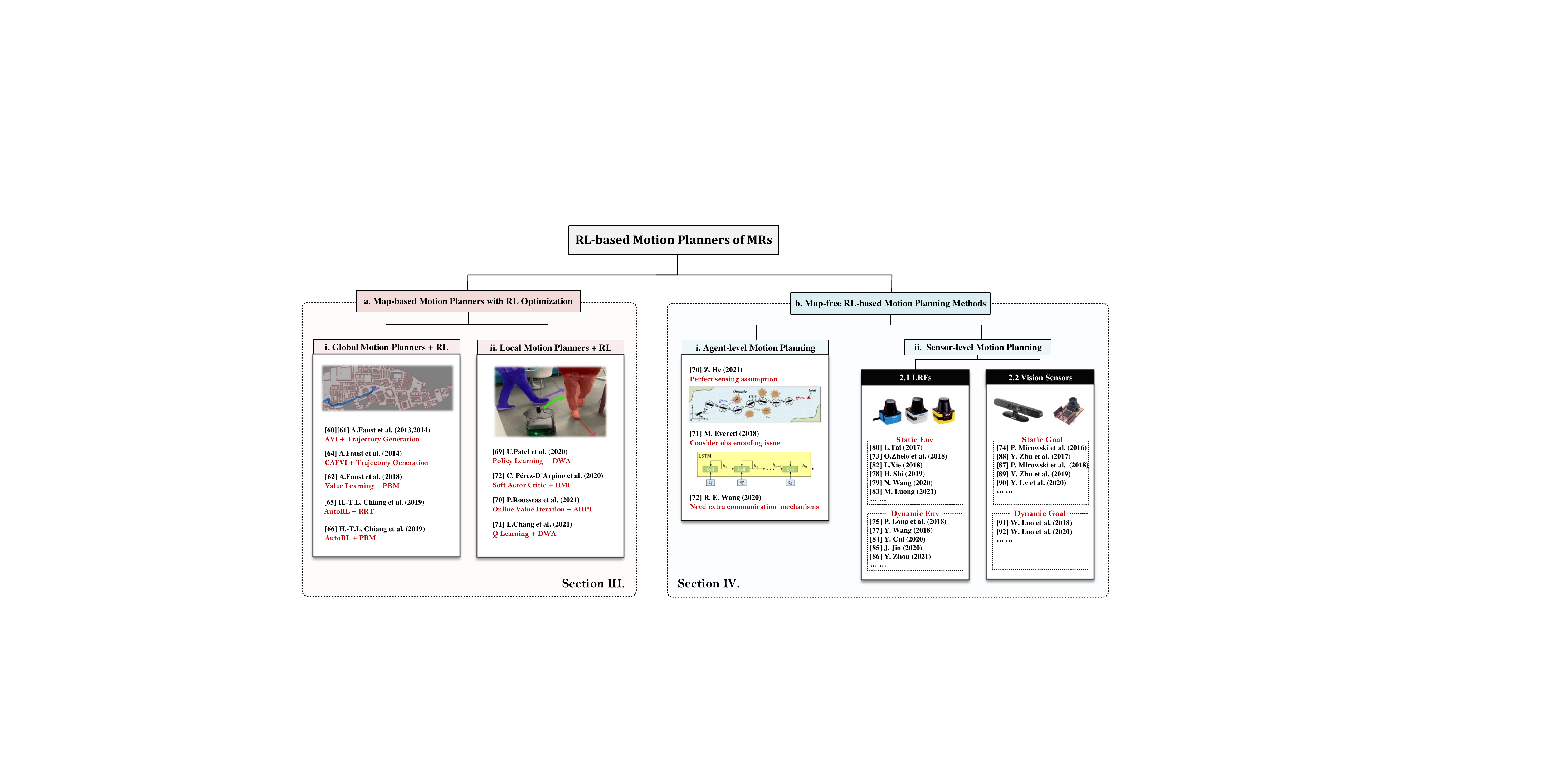}
	\caption{A summary diagram of RL-based motion planning algorithms for the single mobile robot operation. We mainly review two types of research methods: map-based motion planning methods with RL optimization (the left branch, part a.), and the map-free end-to-end RL-based motion planning methods (the right branch, part b.). }
	\label{section_III_IV}
\end{figure*}

\section{Map-based Classical Motion Planning Algorithms with RL Optimization}  

The research of classical motion planning is relatively mature and systematic. However, there are several limitations of these works. For example, the global planner and the local planner of the planning architecture are independent of each other. Researchers need to develop them separately. In addition, to obtain the final executable trajectory, researchers have to solve the optimization problem with several constraints. This process increases the complexity of the planning method. Besides, the classical motion planner contains various sub-modules, each of which requires a tedious parameter adjusting process. As shown in Fig. \ref{section_III_IV} (the left branch), there are considerable works dedicated to combining the advantages of the RL and the classical motion planner to achieve better planning results. This section provides an overview of these classical motion planning methods combined with RL improvements. 
%
%
%
%
%
%
%
%
%
 
\subsection{Global Planners Combined with RL Improvements}   
Some researches are devoted to combining RL algorithms with classical global motion planning methods to address some bottlenecks such as long-distance navigation, dynamic obstacle interaction, etc. 

In \cite{faust2013learning}, Aleksandra Faust et al. design an approximate value iteration (AVI)  RL-based motion planning architecture to help the drone with a suspended load to find a safety trajectory with minimal residual oscillations in the static environment. The results of the simulation and the actual experiment show that their planning method has stronger robustness and better scenario generalization ability. Later, they extend this approach in \cite{faust2013aerial}. They apply the global planner PRM to generate a collision-free path in the global static environment and utilize the RL agent as a local planner to generate the swing-free trajectory for the rotorcraft. This improved task decomposition scheme reduces the task difficulty, narrows the action space of the RL agent, and effectively improves the training speed.

In \cite{8461096}, Aleksandra Faust et al. further propose the PRM-RL algorithm to optimize the indoor long-range navigation problem over complex maps. They utilize PRM to partition an entire planar map into multiple local maps, and deploy the local navigation algorithm on each partial map. Different from previous research routes, the PRM-RL can work in the continuous action space with better control accuracy. The pipeline of PRM-RL consists of three phases. In the first phase, the DDPG agent \cite{silver2014deterministic} or the continuous action fitted value iteration (CAFVI) \cite{faust2014continuous} agent is trained to learn the policy of point-to-point navigation. These pre-trained RL agents play the role of local planners. They can choose the optimal motion commands and transfer these motion strategies to the next phase. In the second phase, the RL-improved PRM can generate the roadmap of the deployment environment. In this global roadmap, the connection between any two configurations is a dynamic feasible trajectory output by the RL agent instead of a straight line. In the last phase, a graph search algorithm is deployed to search the optimal path from the roadmap. In \cite{chiang2019rl}, Aleksandra Faust et al. continue to follow their research route and present RL-RRT. Similar to the previous work, RL-RRT uses a DDPG agent to learn a local collision avoidance policy. The DDPG agent acts as a local planner to generate dynamically feasible trajectories to replace the steer function in the RRT framework. RL-RRT avoids the process of solving the optimal boundary value problem (OBVP) in kino-dynamic RRT algorithms \cite{6631299}. Later, Aleksandra Faust et al. apply AutoRL \cite{chiang2019learning} technology to the previous PRM-RL \cite{francis2020long}. Based on the covariance matrix adaptation evolution strategy, AutoRL can automatically adjust the parameters of different reward terms and the architecture of the policy networks. AutoRL-based PRM-RL avoids the tedious process of hyper-parameters tuning, improves the algorithm accuracy, and makes the motion planner more robust to the sensor noise and the environmental uncertainties.

\subsection{Local Planners Combined with RL Improvements} 

Some researchers integrate RL modules in local planners. Their purpose is to give the robot a stronger ability to operating in unstructed, dynamic, and uncertain environments.

Utsav Patel et al. propose the hybrid DWA-RL motion planning approach \cite{ patel2020dynamically}. They utilize the RL algorithm as the upper-level policy optimizer and adopt DWA as the low-level observation space generator. DWA-RL introduces the benefit of the DWA to perform kino-dynamic feasible planning and uses RL to select the optimal velocity commands to maximize the global returns for complex environments. Panagiotis Rousseas et al. integrate the artificial harmonic potential fields (AHPF) with the RL algorithm \cite{rousseas2021harmonic}. They retain the efficient and real-time features of AHPG, while utilizing the value iteration to improve the planning policy iteratively. AHPG-RL endows the local planner with optimality. Lu Chang et al. propose the Q-learning-based DWA \cite{chang2021reinforcement}. The evaluation function of the classical DWA approach is crucial for the final planning result. In different scenarios, the proportion of each weight of the evaluation function should be different. Q-learning-based DWA uses a Q-learning RL module to auto-tune the weights in the DWA evaluation function at each timestep to improve the optimality of the planner. 

In some works, researchers choose to leverage RL algorithms to help robots better predict behavioral patterns of uncertain factors (like pedestrians) in the surrounding environment. In \cite{perez2020robot}, Pérez-D'Arpino et al. present a soft actor-critic (SAC)-based local planner for constrained indoor navigation problems with pedestrian participation. This hybrid planner is composed of the planning module and the RL module. The planning module is responsible for generating feasible waypoints on the basis of the prior map. These waypoints are utilized as guidance and concatenated with the Lidar data and the goal state as input to the policy network of the SAC agent. The output of the policy network is the speed controller command. The planning module helps the robot  reach the target. The RL module focuses on learning different interaction patterns (slowing down, detouring, going backward, etc.) between robots and pedestrians.

\section{RL-based Mapless Motion Planning Methods}  

In some works reviewed in the previous section, RL is only used as a supplement and improvement module to the classical motion planner. In this section, we shift the focus of the review to the end-to-end RL-based motion planning methods. These motion planners are map-free and realize the unification of the global planner and the local planner. Researchers do not need to construct and maintain a prior geometric map of the operating environment that directly affects the final effect of the motion planning. In addition, compared with some supervised learning-based mapless motion planning methods \cite{8653875,8793889}, RL-based motion planning methods can learn and evolve directly from the interaction data between robots and external environments. This mode avoids the construction process of the complex labeled expert dataset. 

As shown in Fig. \ref{section_III_IV} (the right branch), RL-based mapless motion planning approaches can be further divided into two types: agent-level methods and sensor-level methods. Agent-level methods are based on the preset state estimation process and can directly acquire the upper-level state information of the environment. The agent-level methods are easier to train, which allows the robot to obtain optimal planning strategies faster. The observation of the agent-level method has a lower dimensionality, and contains more useful information. The state of the agent-level methods can be equivalent to the output after feature extraction of the raw sensor data. More importantly, the simulator of agent-level methods are much less difficult to develop. However, such methods rely on the assumption of perfect perception \cite{he2021asynchronous}, or require to consider the observation encoding issue \cite{everett2018motion}, or need extra communication mechanisms to share the state information \cite{wang2020r}. These restrictions limit the scalability and application of agent-level methods. 

Sensor-level methods are end-to-end. These methods directly establish the nonlinear mapping from the raw sensor data to the planning decisions. Although the offline training process is more time-consuming than agent-level methods, sensor-level methods do not rely on the perfect sensing assumption. Since the observation input dimension of sensor data is fixed at each time step, there is no necessity to consider the encoding and representation problems in the dynamic environment. Thus, sensor-level methods have better scalability, scenario generalization, and sim-to-real capabilities than agent-level methods. This section is an overview of these sensor-level and end-to-end RL-based motion planning methods. According to the mainstream research trends, as well as the commonly used robot perception methods, we further divide sensor-level RL motion planning methods into two categories: laser range finder (LRF) based methods and visual-based methods. 

\subsubsection{Laser Range Finder based} 
LRF equipment is widely used in map modeling, mobile robot navigation, autonomous driving, etc. In this section, we summarize some state-of-the-art sensor-level RL-based motion planning approaches with LRF. 

In \cite{zhelo2018curiosity}, Oleksii Zhelo et al. consider motion planning problems of several specific scenarios, including long corridors, dead corners, etc. that are not suitable for RL-based planners to learn the optimal or near-optimal policies. Different from some works that pre-define the reward form of navigation \cite{mirowski2016learning,long2018towards}, researchers introduce the intrinsic curiosity module \cite{pathak2017curiosity} to help RL agent obtain the intrinsic reward. This exploration trick helps the planner acquire better generalization ability in the 2D virtual environment. Likewise, in \cite{shi2019end}, Haobin Shi et al. also introduce the intrinsic curiosity module and present a more general end-to-end motion planner based on the A3C framework with the input of the  sparse Lidar data. They successfully deploy their planner from the physical engine to the realistic mixed scene. Yuanda Wang et al. consider the end-to-end motion planning task in the static virtual environment with dynamic obstacles \cite{wang2018learning}. They decompose the whole motion planning task into an obstacle avoidance subtask and a navigation subtask. The collision-free planning module takes the raw sensor data of LRF as input and outputs a 5-dimensional force vector. It is important to note that the Q network in their planner has two streams: the spatial stream and the temporal stream. The spatial stream deals with the raw sensor data, while the temporal stream processes the difference between two consecutive frames of ranging data. On the contrary, the navigation module is training by a conventional Q network architecture. Experiment results show that this approach could obtain a high-performance motion planner in the 2D dynamic simulation environment. In \cite{wang2020learning}, Ning Wang et al. find those extant methods generally require retraining the RL agent in different motion planning scenarios to reduce the generalization error caused by environmental changes. To overcome this catastrophic forgetting problem, they propose the elastic weight consolidation DDPG (EWC-DDPG)-based motion planning algorithm. EWC-DDPG enables the RL agent to acquire continuous learning capabilities without forgetting previous knowledge. This feature allows the planner to generalize across different scenarios. 

Most of the above motion planners can only be deployed in the numerical simulation space. As we all know, the sim-to-real problem has been hindering the real-world application of those learning-based motion planners. In \cite{tai2017virtual}, asynchronous DDPG (ADDPG)-based motion planning algorithm is applied in mapless navigation of the differential WMR. The action space consists of the velocity and the angular velocity of the robot at each time step. The state space in the training stage includes three parts: (1) the 10-dimensional sparse findings of LRF. (2) the 2-dimensional previous action of the DMR. (3) the 2-dimensional relative position of the goal. A goal-reaching reward is set when the WMR approaches the target point, and a certain penalty is given when the DMR collides with the obstacle. Otherwise, $r_t(s_t,a_t)=C(d_{t-1}-d_t)$. C is a hyper-parameter,  $d_{t-1}-d_t$ is the distance difference between the robot and the target point in adjacent timesteps.The whole motion planner is training in VREP \cite{rohmer2013v} virtual engine and inference in the realistic scenarios. The final results show that the proposed ADDPG-based motion planner is more robust during pedestrian interaction than \textit{Move Base}. Linhai Xie et al. present assisted-DDPG (AsDDPG) for training agents to learn local planning policy in the realistic and static obstacle environment without the prior map \cite{xie2018learning}. This DRL framework integrates DDPG with a classical controller (like a PID controller) to replace the random exploration strategy (e.g., $\epsilon$-greedy). The classical controller can output control policy based on the position error between the current position and the goal. It should be noted that the triggering of this controller is determined by the DQN branch in the whole architecture. Sim-to-real experiments suggest that this trick is able to accelerate and stabilize the training phase effectively. In \cite{luong2021incremental}, Manh Luong et al. present an incremental learning paradigm to address the inefficient training issue in sensor-level RL-based motion planners. The incremental learning phase in the training stage is beneficial to optimize the current policy before loop termination. The researchers utilize sim-to-real technology to verify the performance of their motion planner on the \textit{Gazebo} simulator and the real \textit{Pioneer P3-DX} robot platform. Furthermore, in  \cite{cui2020learning,jin2020mapless,zhou2021r}, researchers expect to endow the MR with social safety awareness while performing end-to-end motion planning tasks. Training robots to safely and carefully interact with pedestrians in the environment instead of simply treating them as static or dynamic obstacles. 

In \cite{cui2020learning}, Yuxiang Cui et al. develop a model-based RL motion planning method with social safety awareness. They first obtain priori data from the interaction process between the robot with the realistic scenarios and utilize this dataset to train a world transition model with pedestrian participation in the self-supervised learning paradigm. Finally, they concatenate the real data with the virtual data generated by the world environment model as the observation and input it into the RL architecture to train the planning policy. In \cite{jin2020mapless} and \cite{zhou2021r}, researchers introduce rectangular social-safety zones for robots and pedestrians, and design the corresponding safety interaction reward term based on these zones. Moreover, they adopt end-to-end RL framework and train the motion planning policy in the mixed and complex environment.

\subsubsection{Vision sensors}

\begin{figure}[!t]
\centering
\includegraphics[width=2.5 in]{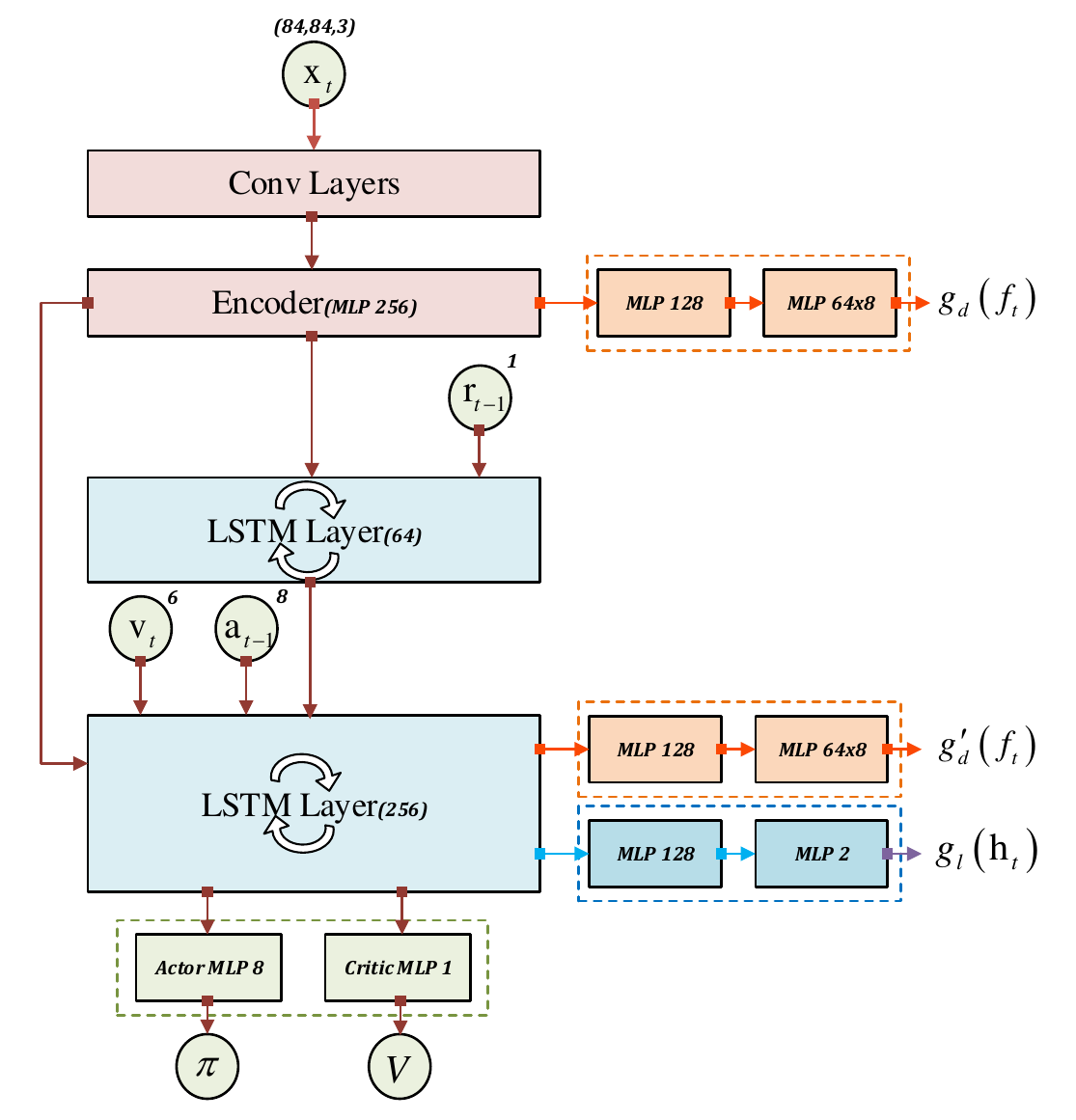}
\caption{The architectures of NavA3C visual navigation method with several auxiliary tasks including depth prediction and closure detection \cite{mirowski2016learning}.}
\label{fig_navA3C}
\end{figure}

In addition to Lidar sensors, vision sensors are also widely used as sensing modules for mobile robots. Many scholars have found that the  end-to-end planning and decision-making for robots can also be achieved based on the raw visual data input. DRL framework combined with convolutional neural networks (CNNs) is naturally suitable for this task. In \cite{mirowski2016learning}, researchers in DeepMind propose the NavA3C algorithm to teach the agent to find the goal and navigate in different 3D mazes. This work encourages robots to learn motion planning policies while learning auxiliary tasks such as environmental loop closure detection and image depth prediction. The whole architecture of NavA3C is shown in Fig. \ref{fig_navA3C}. The inputs contains RGB visual input $\mathrm{\textbf{x}}_t$, past reward $r_{t-1}$, previous action $\mathrm{\textbf{a}}_{t-1}$, and the agent-relative velocity $\mathrm{\textbf{v}}_t$. The outputs including motion policy $\pi$, value function $V$, depth predictions $g_d(\mathrm{\textbf{f}}_t), g_d'(\mathrm{\textbf{h}}_t)$, and closure detection $g_l(\mathrm{\textbf{h}}_t)$. It should be noted that the closed-loop detection and the depth prediction in this paper are based on the supervised learning method. Optimizers need to aggregate all gradients of various loss functions from different tasks in the parameters updating process. Multi-group experiments prove that this trick of adding auxiliary tasks can avoid reward sparsity issues, improve the richness of the training samples, and improve the training efficiency. Later, researchers in DeepMind extend this work, and apply NavA3C to outdoor navigation scenarios on the basis of the realistic street panoramas dataset of Google \cite{mirowski2018learning}. Inspired by NavA3C of DeepMind, Jonáš Kulhánek et al. design auxiliary tasks in the planning architecture to facilitate the domain randomization of the model in simulation environments \cite{kulhanek2021visual}. They directly utilize unlabeled images of segmentation masks that are not readily available in the environment. This operation can significantly improve the planning effect when the planner is deployed in the actual physical environment. 

\begin{figure}[!t]
\centering
\includegraphics[width=3.3 in]{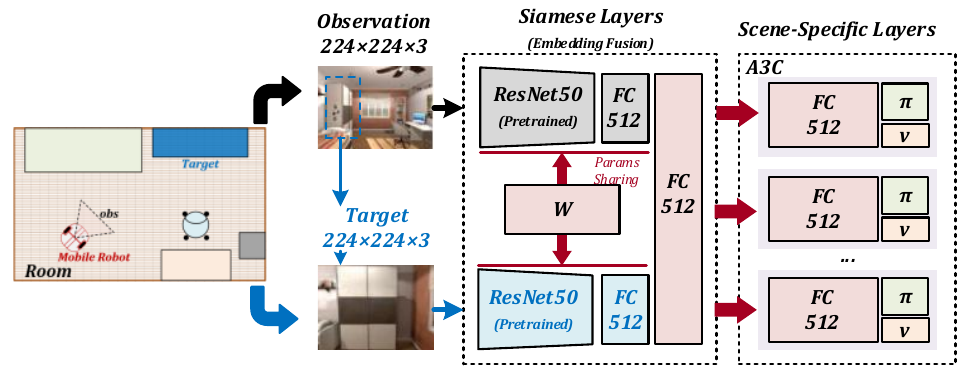}
\caption{The overall architecture of the deep siamese actor-critic based target-driven indoor navigation algorithm \cite{zhu2017target}. This algorithm takes the current observation and the goal image as the input. Two parameter-sharing ResNet-50 are responsible for encoding these two batches of observations, and generating embedding vectors. After fusion operation, the vectors are converted into one embedding vector and input to the A3C as the state input. The A3C scene-specific layers output the final action policy of the mobile robot.}
\label{LifeifeiTargetDriven}
\end{figure}

In \cite{zhu2017target}, Li Fei-Fei et al. propose a DRL-based target-driven visual navigation approach for indoor scenarios. Unlike previous works about visual navigation, their method has a stronger generalization ability to various environments and can be easily deployed to the realistic world just by fine-tuning. The framework of their method is shown in Fig. \ref{LifeifeiTargetDriven}. The inputs include the current observation image and the target image. Weights-sharing siamese networks transfer the input features to the same embedding space. Scene-specific networks based on the A3C output the motion policy and the action value $V$.  Li Fei-Fei and her group have pioneered target-driven visual navigation. They also develop and open-source a high-quality 3D indoor simulation platform: The House Of inteRactions (AI2-THOR) which has subsequently facilitated other scholars to conduct their visual navigation research (e.g., \cite{wu2019exploring,lv2020improving})  
 
%

Above visual-based motion planning methods are based on the premise that the static target is known. Researchers in Peking University and Tencent AI Lab have been jointly working on the visual-based end-to-end motion planning of MRs in active dynamic target tracking scenarios. Their methods utilize the Conv-LSTM network to establish the mapping from the raw sensor image to the control command of the actuator. Visual navigation is generally challenging to achieve the sim-to-real process due to the gap between the simulator and the realistic scene. Aiming at this issue, the researchers perform environment complexity augmentation and virtual model detail augmentation to enhance the robustness and generalization of the algorithm \cite{luo2018end,luo2020end}.  

\section{RL-based Multi-robot Motion Planning}  

There exist several performance limitations in single-robot motion planning operation, such as limited sensing range, low mission reliability, and time inefficiency in some specific sceanrios (e.g., underwater searching, hospital disinfection, etc.). Collaborative motion planning of multiple robots is more flexible, robust, and efficient. Therefore, it is widely applied to marine exploration, smart agriculture, disaster rescue, etc \cite{9119863}.  Unlike the Markov properties of the single-agent RL-based motion planning methods, the RL-based multi-robot motion planning (MRMP) procedure requires consideration of the influence of the local observability and the uncertainty of the environment. Therefore, most of the mainstream works extend the interaction process of robots with the environment from the Markov decision process (MDP) to the partially observable Markov decision process (POMDP) or decentralized POMDP (Dec-POMDP). The major RL-based MRMP research works can be further divided into two main categories: centralized RL-based MRMP (CeRL-MRMP) methods and decentralized RL-based MRMP (DecRL-MRMP) methods. CeRL-MRMP is simple and intuitive. It uses a centralized Q network to build a map from joint trajectories of all agents to the global action-state value and aims at learning the joint planning policy that maximizes the total rewards \cite{tampuu2017multiagent}. CeRL-MRMP methods have some bottlenecks,such as the dimensionality problem of the joint space representation, the exploration problem of the high-dimensional joint policy, and the scalability problem \cite{9119863}. DecRL-MRMP can be further subdivided into the independent MRMP and the centralized training and decentralized execution (CTDE)-based MRMP. Independent MRMP has better scalability \cite{sivanathan2020decentralized}. Each robot only gets partial information of the environment and cannot directly obtain the action policies of other active agents. In independent MRMP, every agent is self-interested and only considers how to maximize own return. Therefore, this type of methods exists credit assignment problem. And for each agent, the entire environment changes dynamically in each timestep, which further leads to the convergence difficulty of the algorithm. 

CTDE-based methods incorporate the advantages of the above paradigms. During the training procedure, each agent can extract global environmental information through a variety of centralized methods, such as total action-state value \cite{ rashid2018qmix }, historical trajectories from the global experience buffer \cite{lowe2017multi}, and other sensor-based explicit approaches like \cite{fan2020distributed}. During the planning execution process, each robot only requires its own local observations to make online inference decisions. In the section, we give an overview of some of these representative works.  

In \cite{lowe2017multi,9119863, yu2021surprising }, researchers focus on improving CTDE-based multi-agent RL (MARL) algorithms. They usually adopt the cooperative navigation scenario of Multi-agent Particle Environment (MPE) \cite{ lowe2017multi } as an experimental benchmark task to test the performance of their proposed MARL algorithms. Specific optimizing objectives of these works include the efficiency of information sharing, the cooperative ability of agents, overall convergence speed, credit assignment strategy, etc. However, agents in MPE navigation environment are too idealized and do not match the characteristics of real MRs. Also, the planner ignores the kinodynamic constraints of the robot in the process of state updating. These limitations hinder the deployment of these algorithms in realistic scenarios.

Researchers in the ACL laboratory of MIT have made great achievements in the field of DecRL-MRMP. Michael Everett et al. are dedicated to studying the problem of MRMP in complex and dynamic scenarios without communication\cite{chen2017decentralized, chen2017socially,everett2018motion,everett2021collision}. They describe this type of problem as a sequential decision-making problem. In a $n$-agent scenario, the state vector of agent $i$ is ${\rm{\textbf{s}}_i}$, and the observation vector of other $n-1$ agents (MRs or pedestrians) is $\widetilde{\mathrm{\textbf{S}}}_i^o$. ${\rm{\textbf{s}}_i=[{\mathrm{\textbf{s}}_i^o,\mathrm{\textbf{s}}_i^h]}}$ where $\mathrm{\textbf{s}}_i^o=[p_x,p_y,v_x,v_y,r]$ represents observable states including the position, velocity and radius of the agent $i$ and $\mathrm{\textbf{s}}_i^h=[p_{gx},p_{gy},v_{pref},\psi]$ represents the unobservable states including the position of goal, the preferred velocity and the orientation of agent $i$. The continuous action space for the agent $i$ at time step $t$ is ${\mathrm{\textbf{u}_t} = [v_t,\psi_t]}$, where $v$ is the speed and $\psi$ is the heading angular. $\pi_i$ is the policy. The objective of each agent i is to develop a policy $\pi^*$ to minimize the time consumption $t_g$ from the start position to the goal with several constraints. The details are as follows. Eqs.(\ref{Mit_ACL2})-(\ref{Mit_ACL4}) respectively represent the safety constraint, the target point constraint and the kinodynamic constraint \cite{everett2021collision}.
\begin{equation}
	\label{Mit_ACL1}{
	\mathop{\mathrm{argmax}}\limits_{\pi_i} \quad \mathbb{E}[t_g | \mathrm{\textbf{s}_i}, \widetilde{\mathrm{\textbf{S}}}_i^o, \pi_i]
	}
\end{equation} 
\begin{equation}
	\label{Mit_ACL2}{
	s.t. \; \left\|{\mathrm{\textbf{p}_{i,t}}-\mathrm{\widetilde{\textbf{p}}_{j,t}}}\right\|_2 \geq r_i+r_j  \quad \forall i\neq j,\; \forall t 
	}
\end{equation}
\begin{equation}
	\label{Mit_ACL3}{
	 \mathrm{\textbf{p}_{i,t_g}}=\mathrm{\textbf{p}_{i,goal}} \quad \forall i \quad\quad\quad\quad\quad
	}
\end{equation} 
\begin{equation}
	\label{Mit_ACL4}{
	\mathrm{\textbf{p}_{i,t}}=\mathrm{\textbf{p}_{i,t-1}} + \Delta t \cdot \pi_i  \quad \forall i \quad\;
	}
\end{equation} 

In \cite{chen2017decentralized},  Michael Everett et al. propose the collision avoidance with deep RL (CADRL) algorithm and apply it to solve (\ref{Mit_ACL1})-(\ref{Mit_ACL4}). They adopt a non-communicating and offline CTDE MARL paradigm to efficiently avoid the time-consuming online computing process in some classical motion planning methods. In the training phase, they utilize a value network $V$ to evaluate the performance of the current policy and iteratively retrieve the optimal time-efficient motion policy from this value function through ${\pi^* = \mathop{\mathrm{argmax}}\limits_{\mathrm{\textbf{u}_t}\in{\mathrm{U}}} \; R([\mathrm{\textbf{s}}_i^o,\mathrm{\textbf{s}}_i^h],\mathrm{\textbf{u}_t})+\gamma^{}V(\hat{s}_{t+1},\hat{\widetilde{S}}_{t+1}^o)}$. CADRL is real-time and has great superiority compared to other mainstream methods. However, the cooperative behaviors of each agent in CADRL-based MRMP method cannot be controlled. In \cite{chen2017socially}, they further extend CADRL and propose the socially aware CADRL (SA-CADRL) algorithm. SA-CADRL introduces social behaviors to the multi-agent motion planning task (The social behaviors here mainly refer to various interacting patterns between pedestrians and MRs). Unlike the existing model-based or learning-based approaches, SA-CADRL integrates the behavior rule of humans (time-efficient rule) and the social norms (passing on the right and overtaking on the left) into the reward function of the RL architecture. Moreover, they deploy the SA-CADRL-based cooperative planner on real MR hardware platform to realize automatic navigation at human-walking speed in pedestrian-rich environments. 
\begin{figure}[!t]
	\centering
	\includegraphics[width=3 in]{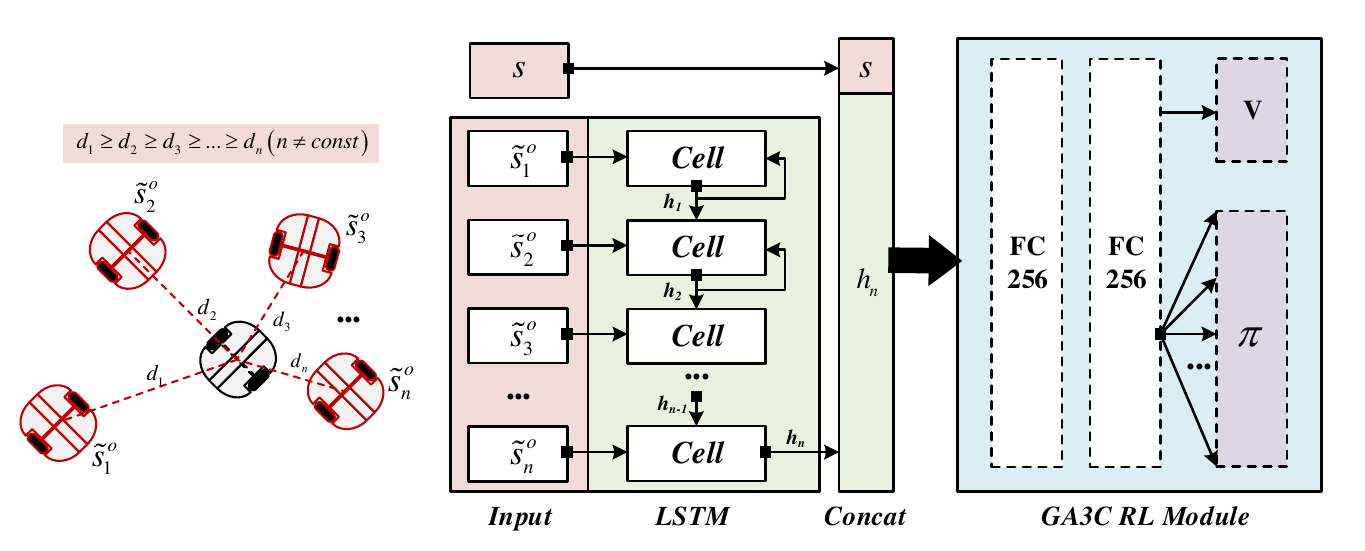}
	\caption{The overall architecture of GA3C-CADRL in \cite{everett2018motion}.} 
	\label{michael_Everett}
\end{figure}
In \cite{everett2018motion}, Michael Everett et al. further consider the stochastic behaviour model and the uncertain number of other agents in the environment on the basis of CADRL and present the GA3C-CADRL MRMP algorithm. The most critical contribution of GA3C-CADRL is that it can tackle the agent-level state representation issue where the number of obstacles or agents in the environment varies randomly. Due to the fixed input dimension constraint of neural network, some researchers choose to define a maximum number of agents and pad the excess space with zero. The utilization of these tricks would introduce several issues like additional parameter calculations and the observation sparsity. Also, compared to those sensor-level MRMP approaches like \cite{tai2017virtual}, Michael Everett et al. hope to extract an agent-level representation that implies the motion plans of other agents. In response to these challenges, they propose an LSTM-based encoding method for environment information and integrate it into the actor-critic RL  framework. The details are shown in Fig. \ref{michael_Everett}. $\mathrm{\textbf{s}}=[\left\|\mathrm{\textbf{p}}_{goal}-\mathrm{\textbf{p}}\right\|_2,v_{pref},\psi,r]$ represents the observation of the agent itself and $\mathrm{\widetilde{\textbf{s}}}^o = [\widetilde{p}_x,\widetilde{p}_y,\widetilde{v}_x,\widetilde{v}_y, \widetilde{r}, \widetilde{r}+r,{\left\|\widetilde{\mathrm{\textbf{p}}}-\mathrm{\textbf{p}}\right\|}_2 ]$ represents the observation of other agents in the vicinity. Whatever the dimension of $\widetilde{\mathrm{\textbf{s}}}^o$ is, the final output $h_n$ can encodes the entire the observation of environment in a fixed-length vector. It is worth noting that Michael Everett et al. open-source their code of simulation environment, which provides a studying platform for other researchers. However, GA3C-CADRL still has certain limitations. First, GA3C-CADRL integrates the supervised learning module. This module relies on the prior dataset and increases the difficulty of training process to some extent. Also, GA3C-CADRL does not improve the reward function in CADRL and still has the reward sparsity issue. To overcome these problems, Michael Everett et al. further improve GA3C-CADRL and propose GA3C-CADRL-NSL \cite{semnani2020multi}. They replace the original navigation reward form with a special goal-distance-based proxy reward function and eliminate the supervised learning stage. The detailed form of this reward function is given in (\ref{MIT_ACL5}).    
\begin{equation}
	\begin{aligned}
		R(s^{jn}) &= R_c + R_g \\ 
		R_c &= \left\{ 
		\begin{array}{lr}
			-1  &\mathrm{if} \ d_{min} < 0 \\
			10d_{min} - 1  &\mathrm{if} \ 0 < d_{min} < 0.1  \\ 
			0  &\mathrm{otherwise}
		\end{array}		
	\right. \\ 
	R_g &= \left\{
		\begin{array}{lr}
			1  &\mathrm{if} \ p = p_g  \\ 
			\propto( \mathrm{goal}^{t-1}_{\mathrm{dist}}-\mathrm{goal}^{t}_{\mathrm{dist}}) &\mathrm{otherwise}
		\end{array}
	\right.
	\end{aligned} \label{MIT_ACL5}
\end{equation} 
where $R_c$ is responsible for monitoring $d_{min}$ which represents the distance between the current agent $i$ and its closest agent, and punishing dangerous actions. $R_g$ is responsible for rewarding goal-reaching discrete action pairs. This reward shaping trick can generate continuous reward signals. It should also be noted that GA3C-CADRL-NSL introduces a hybrid motion planning architecture that combines DRL and force-based motion planning(FMP) \cite{semnani2020force} method. Once the mobile robot falls into high-risk situations, the FMP algorithm will take effect and help the robot get out of trouble.  

\begin{figure}[!t]
	\centering
	\includegraphics[width=3.1 in]{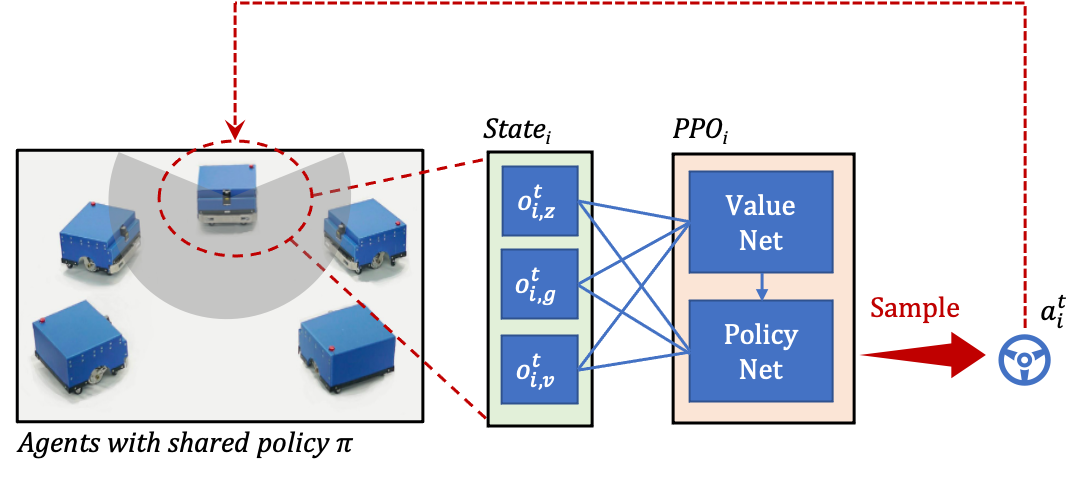}
	\caption{The overall framework of the distributed PPO-based multi-scale mobile robots motion planning algorithm in \cite{long2018towards}.} 
	\label{HongKong1}
\end{figure}

\begin{figure}[!t]
	\centering
	\includegraphics[width=3.1 in]{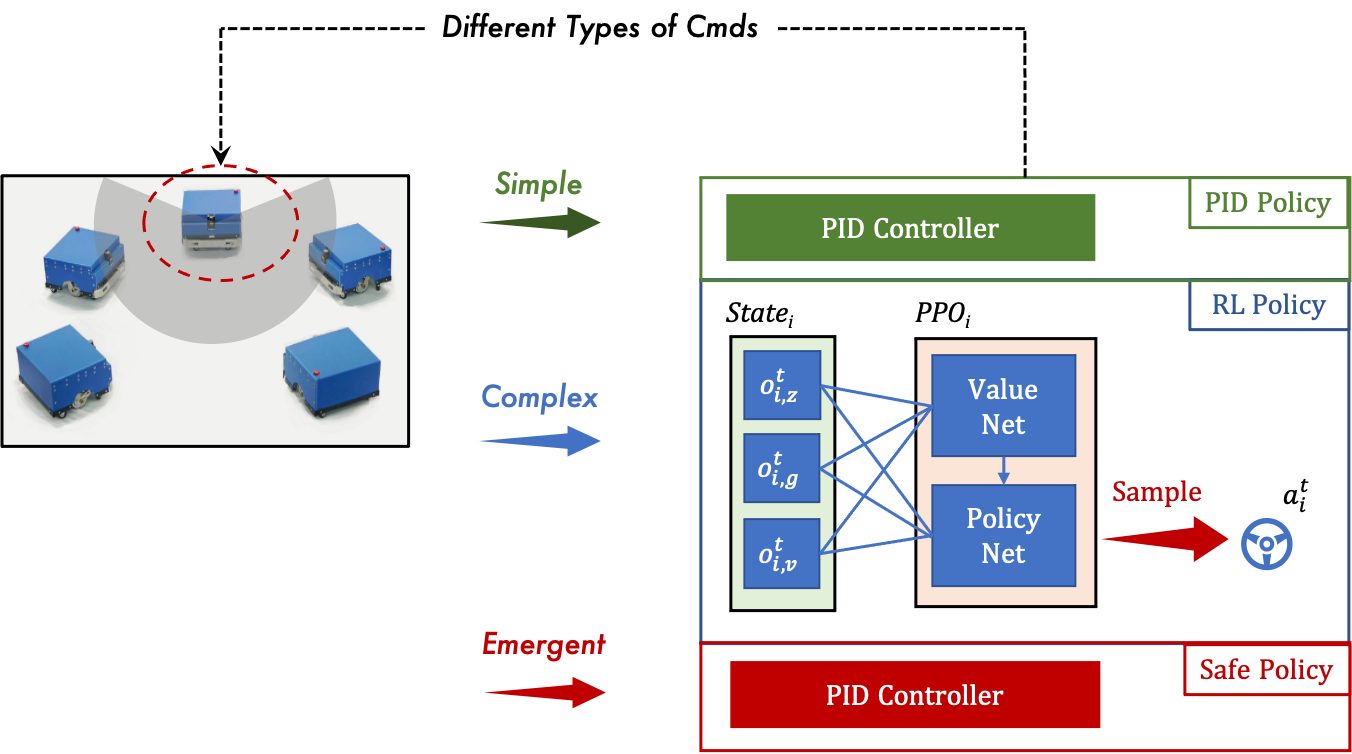}
	\caption{The hybrid motion planning framework in \cite{fan2020distributed}. The robot will choose an appropriate action command according to the type of the current external environment.}
	\label{HongKong2}
\end{figure}

Jia Pan et al. have been working on DecRL-MRMP for large-scale multi-robots in dense environments \cite{long2018towards}. Different from previous works  \cite{everett2018motion}, they pay more attention to the sensor-level motion planning methods. They argue that the assumptions underlying the agent-level approaches are too strong and not general. In addition, they think that agent-level methods have to cooperate with environmental encoding methods to obtain the fixed-dimensional observation input in a dynamic environment \cite{chen2017decentralized}. In their CTDE-based MRMP approach, each robot is independent of others, which makes their method has strong scalability \cite{tang2018hold}. During the training phase, rewards, policy networks, and value networks are shared among each robot. Also, the shared transition samples  are utilized to guide the development of implicit collaboration mechanisms. The main idea of \cite{long2018towards} is shown in Fig. \ref{HongKong1}. Later, they combine this method with the PID controller and propose a hybrid architecture that can be deployed in real-world scenarios \cite{fan2020distributed}. The detail is shown in Fig. \ref{HongKong2}.

\section{Discussion}
%
%

\begin{figure*}[!t]
	\centering
	\includegraphics[width=6.7 in]{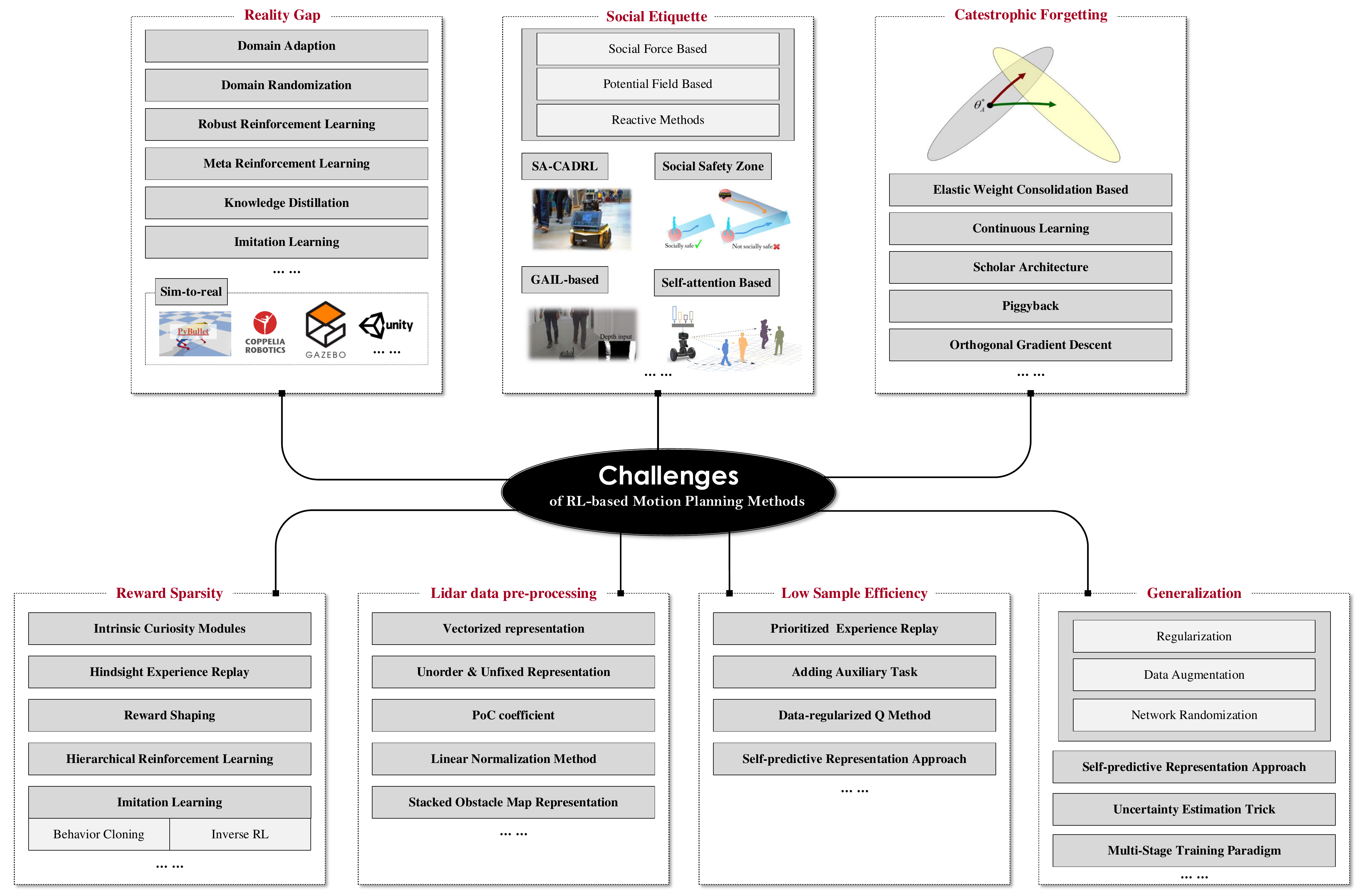}
	\caption{ A summary diagram of several issues existing in RL-based  motion planning methods. Current RL-based motion planning approaches have many performance limitations, such as reality gap, reward sparsity, generalization, low sample efficiency, social etiquette, lidar data pre-processing issue, catastrophic forgetting problem.}
	\label{Challenges}
\end{figure*}  


At present, RL-based motion planners still face plenty of challenges. These challenges stem from the characteristics of the algorithm itself, the limitations of sensors, and the external environment. Therefore, many factors still need to be considered before realizing the large-scale deployment of the RL motion planner to real navigation scenarios. In this section, we mainly analyze and summarize these challenges. The specific contents are summarized in Fig. \ref{Challenges}. Then, in conjunction with these analyses, we provide suggestions for future research directions. 

\subsection{Challenges}  
\subsubsection{Reality Gap} 
There is a real-world gap in applying DRL to realistic robotic tasks. For instance, unexpected actions may cause potential safety problems (a real robot could cause real damage) of robots in real-world scenes, low sample efficiency in the real world may lead to convergence difficulty of the training process. Besides, sensors and actuators of real robots cannot be as ideal as virtual environments, which brings plenty of uncertainties. At present, many scholars in Robotics have committed to the research of innovative Sim-to-Real methods \cite{zhao2020sim}. Mainstream Sim-to-Real approaches include domain adaption methods \cite{zhang2019vr}, disturbances learning-based robust methods \cite{lutjens2019safe}, domain randomization methods \cite{chaffre2020sim,kulhanek2021visual}, knowledge distillation methods \cite{kalifou2019continual}, etc. 

As for motion planning tasks, training planning policy in the simulation platforms with the physical engine (Some popular platforms include CARLA, Pybullet, CoppeliaSim, Gazebo, Unity 3D, etc.) and transferring to the real-world navigation scenario is a commonly used research pipeline to alleviate the influence of the reality gap problem \cite{rusu2017sim,zhu2019learning}. For example, Thomas Chaffre et al. present a depth-map-based Sim-to-Real robot navigation method. They first set up several scenarios with increasing complexity in the Gazebo platform for incremental training. In the real-world scenario training stage, the learning phase is deployed on the fixed ground truth Octomap and utilizes a PRM* path planner to ensure safety in the resetting stage of every episode. The velocity and the angular velocity commands are output to the low-level controller of the MR in the testing phase \cite{chaffre2020sim}. Jingwei Zhang et al. handle the Sim-to-Real motion planning problem by adapting the real camera streams to the synthetic modality during the actual deployment stages. This operation is lightweight and flexible, and could transfer the style of realistic image to the simulated style, which can be used in the stage of training RL agent \cite{zhang2019vr}. Similarly, Jing Liang et al. propose a brand new learning-based local navigation approach named CrowdSteer to solve the motion planning problem of MRs in dense environments \cite{liang2020realtime}.


\subsubsection{Sparse reward problem}
The motion planning process of mobile robots is often target-driven. The positive reward of the environmental feedback is generally at the final goal point. For long-distance navigation in obstacle environments, it is hard for robots to obtain final positive reward signals. Besides, long-term training of robots under the negative reward signals might develop abnormal behavior patterns
, such as timidity. Moreover, the sparse reward problem can lead to slow learning problems and convergence difficulties.

Curiosity-driven is a way to solve the problem of sparse rewards using existing trajectories \cite{pathak2017curiosity}. The main idea of this method is to build intrinsic curiosity modules (ICM) to extract additional intrinsic reward signals from the environment to encourage more effective exploration of the agent. In \cite{zhelo2018curiosity,shi2019end}, researchers choose to utilize this trick to develop map-free and end-to-end motion planning frameworks of MRs. Hindsight experience replay (HER) \cite{andrychowicz2017hindsight} is another approach used to solve the reward sparsity problem with existing data. HER is based on the multi-objective RL algorithms. The main idea of this algorithm is to encourage learning from unrewarded states. By mapping the unrewarded state as the new target and replacing the previous target, the agent is encouraged to explore and obtain additional reward signals during the training process. Different from above methods, reward shaping is more intuitive. Researchers using this trick must be patient and manually adjust and modify the refined reward signal values of the robot under different states\cite{sun2019mapless}. So, reward shaping skill is highly dependent on expert experience. An improper reward can lead to the change of  the optimal policy and cause the anomalous behaviors of the agent\cite{DBLP:conf/icml/NgHR99}. Another type of method to solve reward sparsity is the hierarchical RL (HRL). HRL tends to decompose the original task hierarchically into multiple discrete or continuous and easy-to-solve subtasks, and then divide and conquer to provide the agent with dense reward signals. Besides, there are already some researchers studying HRL-based motion planners, such as \cite{qiao2020behavior, christen2021learning}. For some more specific and complex planning tasks, the reward functions are hard to configure. Expert demonstrations can be utilized to help mobile robots learn better, i.e., imitation learning. Common imitation learning paradigms include behavior cloning and inverse RL (IRL). Behavior cloning relies on a supervised learning process and suffers from the mismatch problem between the actor and the expert policy. In \cite{chi2019collision}, Zijing Chi et al. utilize this method to help the robot develop the collision avoidance ability. In contrast, IRL is not just a simple imitation of expert behaviors. IRL makes good use of the expert trajectories to learn the reward function in the RL architecture in reverse and performs the policy optimization after obtaining the reward function. IRL methods are commonly applied to the autonomous driving field \cite{you2019advanced,rosbach2019driving}. 

\subsubsection{Generalization} 
The generalization ability of the RL-based motion planners determines whether the mobile robot can perform safe and reasonable motion primitives in unseen scenarios different from the training stage. Most of the RL-based motion planners rely heavily on the inference performance of neural networks. However, for the unpredictable far-from-training test cases or the out-of-distribution test data, neural networks cannot guarantee the security and effectiveness of the planning process \cite{lutjens2019safe}. For example, cleaning robots have to learn to interact with pedestrians in a school campus. If the robot cannot correctly predict the movement intention of passers-by, there will be some potential safety hazards. At the algorithm level, many scholars design improving methods to enhance the generalization ability of RL agents, such as the regularization method \cite{cobbe2019quantifying}, the data augmentation trick \cite{laskin2020reinforcement}, etc. In \cite{lee2019network}, Kimin Lee et al. propose a method to improve the generalization ability of the RL agent across different tasks by using random neural networks to generate random observations. In addition, some generalization enhancing techniques have been applied to the RL-based motion planning methods. In \cite{fan2020distributed}, Tingxiang Fan et al. adopt the multi-stage stochastic training policy to improve the generalization of the MR in different environments. In practical applications, the pre-trained RL policy is combined with a PID policy and a safe policy to ensure the stability and safety of the robot in unseen scenarios. In \cite{lutjens2019safe}, a safe RL architecture is proposed to handle dynamic collision avoidance problems in novel scenarios. Unlike the traditional RL framework, this method integrates the collision prediction networks based on the LSTM ensemble, the uncertainty estimation based on Monte-Carlo dropout, bootstrapping process, and the safest action selection method based on the model predictive control (MPC). The final results demonstrate that this type of uncertainty-aware pipeline endows the motion planner with stronger robustness and generalization. 
  
%

\subsubsection{Low sample efficiency}
Unlike the manner of human learning, the RL agent learns from scratch for different tasks. In the early stage of training, too much useless transitions make it difficult for the agent to learn and update effectively. Therefore, how to better explore valuable strategies and enhance the sampling efficiency of valuable experience are the hot directions in the RL-related research, and they are also the key to improving the performance of RL-based motion planners. 


Schaul et al. propose prioritized experience replay (PER) method \cite{schaul2015prioritized}. In order to improve the learning efficiency, Schaul et al. compute the probability of a sample according to the importance (i.e., temporal-difference error). PER is indeed an effective trick to improve the efficiency of off-policy RL methods. There is already some work applying PER to RL-based motion planning methods \cite{zijian2021relevant,9327315}. Lasjub et al. utilize constractive unsupervised representations as the auxiliary task to speed up sample efficiency. \cite{laskin2020curl}. They extract valid features from the raw sensor data through the comparative learning process and then feed the features into the RL module. Kostrikov et al. proposed Data-regularized Q (DrQ) method \cite{kostrikov2020image}. They do augmentation on the observation input before starting the sampling training process and calculating the target-Q and the current Q simultaneously. Also, through combining with the regularization trick, the sample efficiency of the raw input data is significantly improved. Schwarzer et al. propose the self-predictive representation (SPR) approach \cite{schwarzer2020data}. They improve the sample efficiency by training agents to predict multi-step representations of their potential future states. This operation allows agents to learn temporally predictive and consistent representations under different environmental observations. 

\subsubsection{Social etiquette} 
At present, more and more robots are deployed in crowded places such as airports and stations. In these specific scenarios, mobile robots using classical motion planners may cause the freezing problem (Robots cannot find any feasible action) because the probabilistic evolution of pedestrians could expand to the entire workspace \cite{trautman2010unfreezing}. Therefore, it is challenging but essential to deploy learning-based algorithms to train MRs to learn social etiquette and interact with humans in a safe, effective, and socially compliant manner. 

Pioneer research works including social force-based methods \cite{ferrer2013robot}, potential field-based methods \cite{ferrer2014behavior}, reactive methods (RVO, ORCA) \cite{van2008reciprocal,van2011reciprocal}, etc. These methods are overly dependent on the hand-crafted process and lack a certain generalization ability for complex scenarios. Some works simplify the motion pattern of pedestrians. They treat pedestrians as static obstacles or dynamic obstacles with simple kinematics over short timescales \cite{phillips2011sipp,chen2017decentralized}. The effectiveness of these approaches is based on several strong assumptions. When the planner is deployed to the real dynamic environment, robots may produce unsafe decisions due to their inaccurate prediction of human behaviors. 

Typical RL-based methods including SA-CADRL \cite{chen2017socially}, GAIL-based planning methods \cite{tai2018socially}, etc. These methods effectively constrain the specific interaction norm between the robot and the pedestrian, but rely on effective explicit pedestrian detection approaches. In \cite{jin2020mapless} and \cite{zhou2021r}, researchers construct rectangular social-safety zones for the MR and pedestrians respectively, and design corresponding safety reward terms. In \cite{chen2019crowd}, Changan Chen et al. propose a self-attention and deep V learning-based agent-level crowd-aware robot motion planning approach. They consider more practical crowd-robot interaction patterns rather than the first-order human-robot interaction pattern problem. The state value estimation network of this deep V learning framework consists of three modules: an interaction module, a pooling module, and a planning module. The interaction module has a multi-layer perception (MLP) to extract the pairwise interaction feature between the robot and the nearby pedestrians. The pooling module outputs embedding tensors of above pairwise interactions by self-attention model. The final planning module estimates the state value based on previous compact embeddings. 

\subsubsection{Lidar data pre-processing issues} 
Lidar data represents the distance information between the MR and the surrounding environments. Compared with the visual sensors, Lidar data naturally contains depth information and is much easier to achieve the sim-to-real process. Therefore, Lidar is widely utilized in end-to-end motion planning tasks. However, improper Lidar data pre-processing may cause the degradation of the planning capability of the robot in unknown scenarios.

Many works have directly utilized the distance vector read from Lidar as part of the observation in the RL framework (e.g., \cite{tai2017virtual,xie2020learning}). This operation may cause some issues. For example, if the Lidar observation at a certain timestep in the testing scenario is similar to the training scenario but with different passability, the agent may not make different action decisions. Also, if the Lidar data occupies most of the dimensional space in the observation input, the agent may not have a good goal-reaching ability for planning in obstacle-free scenarios. Francisco Leiva et al. propose an unordered Lidar data representation method with the non-fixed dimension \cite{9145667}. This method integrates the relative distance and the orientation information of obstacles, making the whole motion planning algorithm more robust. Wei Zhang et al. present a Lidar data preprocessing approach with the parameter self-learning mechanism \cite{zhang2021enhancing}. They introduce the PoC (proportion of distance values considered "close") ratio coefficient to differentiate similar scenarios and help the MR judge the complexity of the surrounding environment.  Yuxiang Cui et al. in \cite{cui2020learning} find that the stacked obstacle map generated based on the 2D laser scan data has a lower reconstruction error and can represent the difference between the static and dynamic obstacles in the environment more accurately than the angle range representation method.

\subsubsection{Catastrophic forgetting problem}
The catastrophic forgetting problem of RL-based motion planners refers to the forgetting of previously learned knowledge by agents when performing task-to-task continuous learning processes. Since the motion planning process of robots generally involves multiple optimization objectives, the weights that are important for previous tasks might be changed to adapt to a new task. If changes contain highly relevant parameters to historical information, the new knowledge will overwrite the old knowledge, resulting in catastrophic forgetting issues. Kirkpatrick et al. propose the elastic weight consolidation algorithm \cite{kirkpatrick2017overcoming}. They solve the forgetting problem by calculating the Fisher information matrix to quantify the importance of the network parameter weights to the previous task. Next, they add this term as a regularization to constrain the update direction of the neural network while learning a new task. In \cite{wang2020learning}, Ning Wang et al. combine EWC with DDPG algorithm and apply it to the multiple target motion planning task of the mobile robot. In \cite{shin2017continual}, Shin et al. propose a scholar architecture with a generator and a solver. The old generator generates replay data and mixes it with the current task data as the training sets for the new task. This operation ensures that the new scholar does not forget the previous knowledge while learning a new task. Mallya et al. present Piggyback \cite{mallya2018piggyback}. Researchers fix a backbone network and train a binary mask network for each task. Different binary masks are combined with the backbone network to perform different policies to simplify the computation process and improve reusability. Farajtabar et al. propose the orthogonal gradient descent method \cite{farajtabar2020orthogonal}. This approach reduces the forgetting problem of existing knowledge by orthogonally projecting the updated gradients of the new task on the gradient parameter space of the previous task.

\subsection{Future Directions}  

\subsubsection{Task-free RL-based general motion planner }
A complete motion planning task generally consists of several subtask goals. The main idea of the commonly used continuous learning approach is to learn task by task and to overcome the problem of forgetting during task transferring. These operations cause the training phase to be multi-stage and cumbersome. Combining the multi-task motion planning process with the state-of-the-art task-free continuous learning paradigm can directly determine the state of the model based on the fluctuation information of the loss function. It facilitates breaking the rigid boundary between individual subtasks and trains the general motion planner that accomplishes multiple task goals. 

\subsubsection{Meta RL-based motion planning methods} 
Meta learning helps the model to learn how to learn by acquiring sufficient prior knowledge in a large number of tasks. In the process of training the RL-based motion planner, the meta learning mechanism can be introduced to inspire the robot to learn to inference in unknown environments. Meta RL can make the robot adapt to the planning task in the new environment quickly by using the prior knowledge gained from the previous planning experience, and improve the generalization ability of the motion  planners.

\subsubsection{Multi-modal fusion based RL motion planning methods}
Multi-modality consists of two levels: the sensor level and the data representation level. At the sensor level, the utilization of single-type perception sensors has performance limitations. By combining mapless end-to-end motion planning methods with multi-sensor fusion techniques, the advantageous features of each perception module can be fully leveraged and complemented. Moreover, this fusion trick improves the environment cognition and understanding ability of robots and enhances the fault tolerance and robustness of motion planners. At the data representation level, multi-modal RL-based motion planning refers to integrating data features from multiple modalities (e.g., images, languages, etc.) during the training process. For instance, it is possible to improve the planning performance through integrating human language instructions with the visual information as the observation to improve the motion planning performance of the robot for real deployment applications.  

\subsubsection{Multi-task objectives based RL motion planning methods}
A practical motion planner is often developed with multiple task objectives. For example, researchers usually expect the planner to guarantee the shortest path length and the safety of the MR while having as little time and energy consumption as possible during the whole process. Classical motion planner separates the planning process and the optimizing process. Therefore, it is worthwhile to research how to introduce multi-task objectives learning in map-less end-to-end planning architecture. The breakthrough point is to improve the generalization of the planning methods by treating the domain information contained in the training signals of related tasks as induction bias. This operation helps the algorithm learn general skills that could be shared and utilized across various related tasks, and maintain a competitive balance between multi-task objectives. Finally, The representation of the planning policy can be obtained by merging operation multiple task-specific policies into a unified single optimal policy. 

\subsubsection{Human-Machine interaction mode based motion planning methods} 
Most of the current planning application scenarios for RL-based motion planners are point-to-point. The robot learns independently throughout the planning loop. In more complex and changing unstructured environments, human intelligence can be coupled with machine intelligence. A prevalent human-in-the-loop approach is to endow the human with the role of supervisor. The robot autonomously performs the human-assigned task for a period of time, then stops and waits for the next cycle of planning commands. This approach makes the MR unable to respond effectively to sudden external changes. Therefore, it is necessary to study how to integrate human and robot intelligence to improve the human-machine interaction motion planning performance in-depth, such as teaching by demonstration, learning based on human judgment and experience, etc. 

\subsubsection{RL-based motion planning of multiple heterogeneous MRs}
Most of the multi-robot RL-based motion planning approaches in this survey work in 2D space, and each robot of the system shares the same action space and the same observation space. The multiple heterogeneous MRs system has a more extensive application domain (e.g., air-ground cooperation planning and air-sea cooperation planning, etc.). It can leverage the unique advantages of each single-structured robot in the system. On this basis, the advantages of the centralized critic and distributed actors architecture of the MARL could be utilized to realize the dynamic task allocation of each heterogeneous robot and achieve optimal joint planning decisions in 3D space without the prior map. 

\subsubsection{Multi-MR flexible formation planning methods}
Multiple MR swarming and formation planning is widely used in military, logistics, transportation, intelligent agriculture, and resource exploration, etc. To realize the combination of the mainstream RL-based motion planner with the flexible formation, researchers can utilize the hierarchical learning paradigm and continuous learning paradigm or design multimodal reward function to implement the overall planning of the robots group and the formation-keeping within the robots group. In addition, another breakthrough point is to design a hybrid planning framework that incorporates RL-based motion planners and the mainstream formation algorithm to enhance the formation and planning performance.  

\section{Conclusion}  
In this paper, we systematically review the state-of-the-art motion planning methods of mobile robots and give an overview of RL-based motion planners. There are three mainstream research directions: motion planners combined with RL improvements, map-free RL-based motion planning methods, and RL-based multi-robot cooperative planning methods.  Although there are many representative research works, RL-based motion planners still have several performance bottlenecks that hinder its pratical application, such as reality gap, reward sparsity problem, low sample efficiency, generalization problem, catestrophic forgetting problem, social etiquette, Lidar data pre-processing issue, etc. At last, we analyze these challenges and predict the future directions of RL-based motion planning methods. 


%

%
%
%

%

\ifCLASSOPTIONcaptionsoff
  \newpage
\fi



%
%
%

\bibliographystyle{IEEEtran}
\bibliography{References.bib}

%


%
%




\end{document}